\documentclass[10pt,twocolumn,letterpaper]{article}

\usepackage[pagenumbers]{iccv} 

\definecolor{iccvblue}{rgb}{0.21,0.49,0.74}
\usepackage[pagebackref,breaklinks,colorlinks,allcolors=iccvblue]{hyperref}

\usepackage{algorithm}
\usepackage{algorithmic}
\usepackage{amsfonts}
\usepackage{amssymb}
\usepackage{booktabs}
\usepackage{amsmath}
\usepackage{graphicx}
\usepackage{tabularx} 
\usepackage{array}
\usepackage{adjustbox}

\usepackage{colortbl}
\usepackage{makecell}
\usepackage{multirow}

\def\paperID{13875} 
\def\confName{ICCV}
\def\confYear{2025}

\title{iConFormer: Dynamic Parameter-Efficient Tuning \\with Input-Conditioned Adaptation}

\author{
\textbf{Hayeon Jo}\textsuperscript{1},\ 
\textbf{Hyesong Choi}\textsuperscript{1},\ 
\textbf{Minhee Cho}\textsuperscript{1},\ 
\textbf{Dongbo Min}\textsuperscript{1,*}\\[0.5em]
\textsuperscript{1}Ewha W. University
}

\begin{document}
\maketitle
\begin{abstract}
Transfer learning based on full fine-tuning (FFT) of the pre-trained encoder and task-specific decoder becomes increasingly complex as deep models grow exponentially. Parameter efficient fine-tuning (PEFT) approaches using adapters consisting of small learnable layers have emerged as an alternative to FFT, achieving comparable performance while maintaining high training efficiency.
However, the inflexibility of the adapter with respect to input instances limits its capability of learning task-specific information in diverse downstream tasks. In this paper, we propose a novel PEFT approach, \textbf{i}nput-\textbf{Con}ditioned trans\textbf{Former}, termed \textbf{iConFormer}, that leverages a dynamic adapter conditioned on the input instances. To secure flexible learning ability on input instances in various downstream tasks, we introduce an input-Conditioned Network (iCoN) in the dynamic adapter that enables instance-level feature transformation. To be specific, iCoN generates channel-wise convolutional kernels for each feature and transform it 
using adaptive convolution process to effectively capture 
task-specific details tailored to downstream tasks. Experimental results demonstrate that by tuning just 1.6\% to 2.8\% of the Transformer backbone parameters, iConFormer achieves a performance comparable to FFT in monocular depth estimation and semantic segmentation, while outperforming it in image classification and instance segmentation. Additionally, the proposed method consistently outperforms recent PEFT methods for all the tasks mentioned above.
\end{abstract}    
\begin{figure}[t]
\centering
\includegraphics[width=0.98\columnwidth,height=0.35 \columnwidth]
{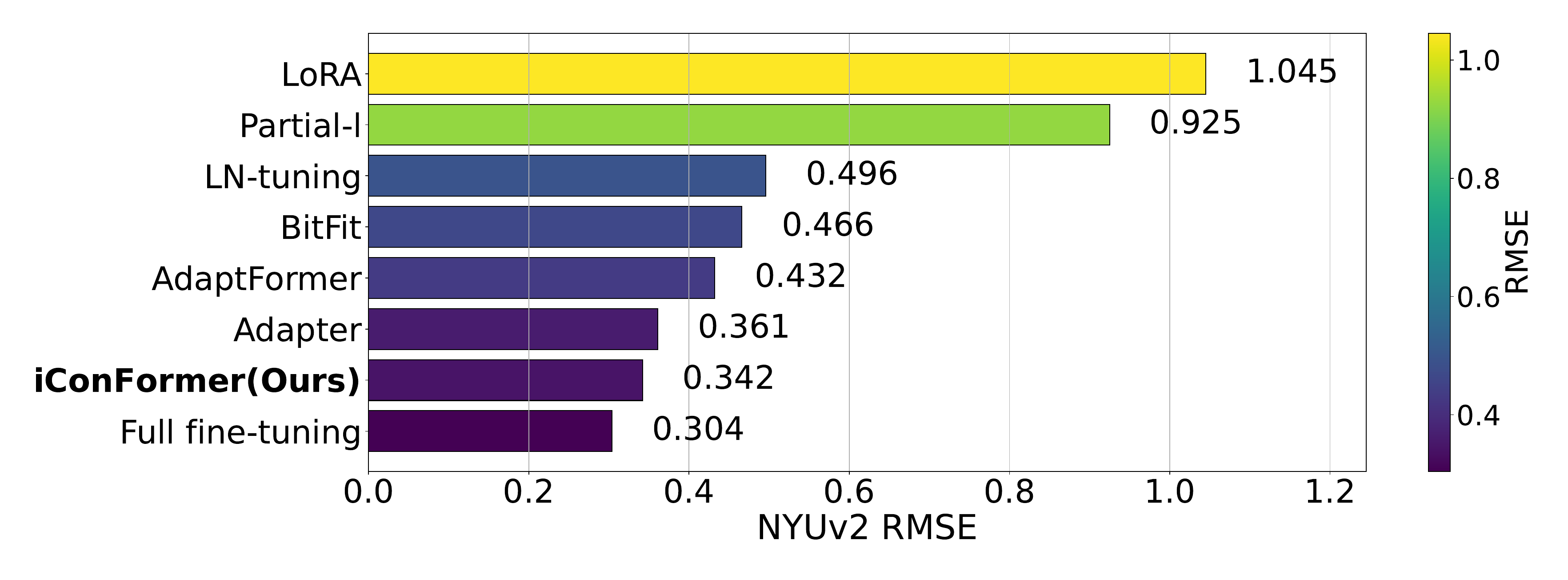} 

\centering
\includegraphics[width=1.0\columnwidth,height=0.65 \columnwidth]
{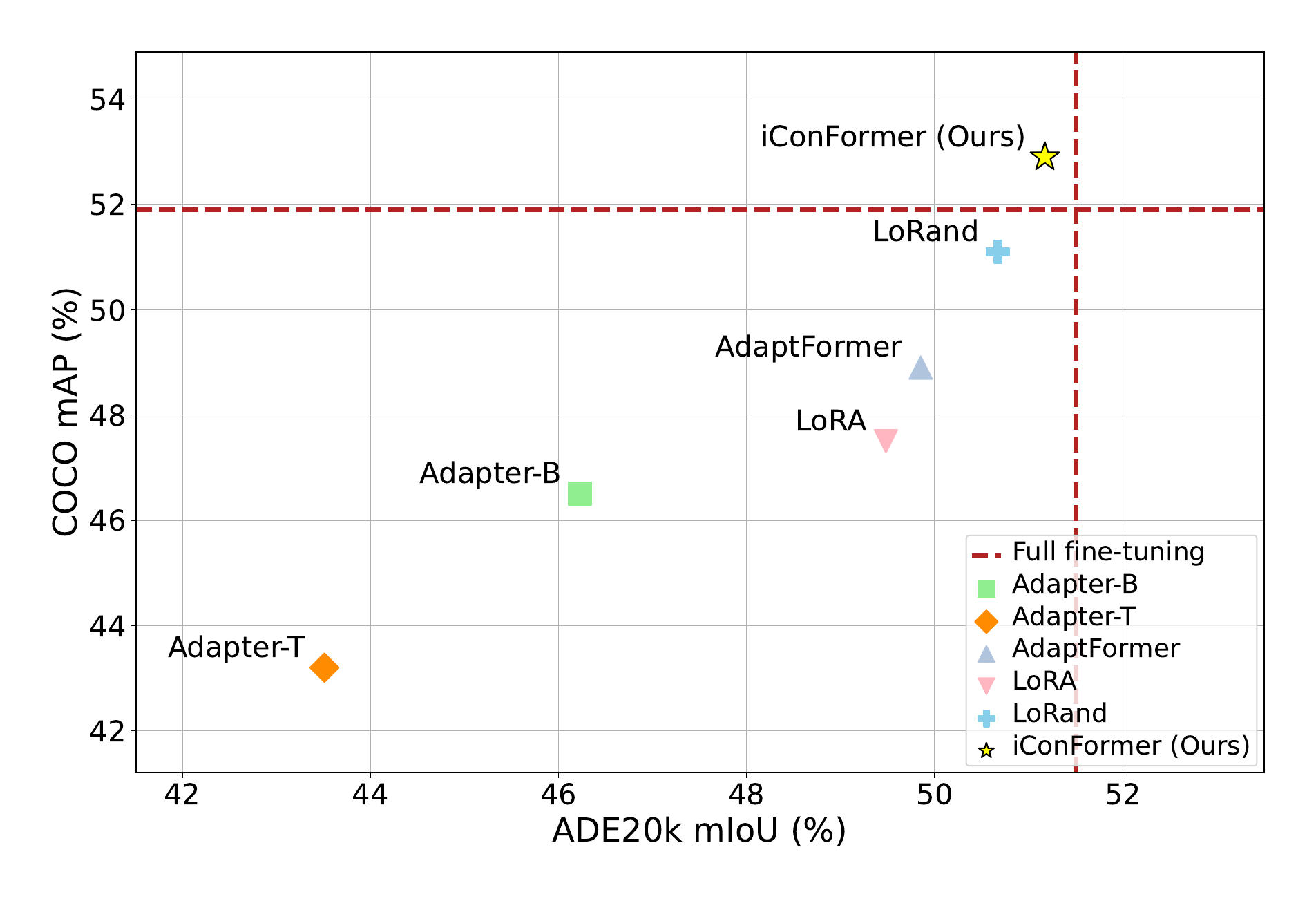} 
\vspace{-25pt}
\caption{\textbf{Quantitative comparison with full fine-tuning (FFT) and PEFT approaches.} The top graph compares depth prediction errors on NYU-v2 \cite{nyuv2}, while the bottom graph presents performance for semantic segmentation on ADE20K \cite{ade20k} and instance segmentation on COCO \cite{coco}. iConFormer consistently outperforms recent PEFT methods in all dense tasks and surpasses FFT in instance segmentation.} 
\label{analysis_graph}
\end{figure}

\section{Introduction}
\label{sec:intro}

As deep neural networks (DNNs) grow increasingly complex, transfer learning—fine-tuning pre-trained models with task-specific data for downstream tasks—has become a widely adopted solution across diverse applications, including image classification, semantic segmentation, and object detection, to name a few. For instance, the model consisting of the pre-trained encoder~\cite{simmim, mae} and task-specific decoder is fine-tuned, achieving remarkable performance gain when compared to the model trained from scratch. However, training large, complex models separately for each task and dataset is highly inefficient.
Recently, parameter efficient fine-tuning (PEFT) approaches~\cite{bitfit, LN-tuning, lora} that maximize the efficiency in terms of training parameters have emerged as an alternative to the above-mentioned full fine tuning (FFT) methodologies, achieving competitive performance even with limited computing resources while simplifying the training processes and deployment.

This remarkable progress in vision tasks is primarily driven by approaches including prompt tuning~\cite{prompt_nlp} and adapter~\cite{adapter_nlp}, which have been successfully applied to natural language processing (NLP) tasks. Visual Prompt Tuning (VPT)~\cite{vpt} is the first study to explore the potential of prompt tuning in visual recognition tasks, laying the foundation for the prompt tuning in the field of computer vision. In addition, the adapter-based PEFT methods~\cite{adaptformer, lorand} achieve significant training efficiency by applying the adapter to the Vision Transformer (ViT) and its variants~\cite{vit, swin, swinv2}.

While most PEFT-based approaches yield performance comparable to baseline methods using the FFT in the image classification task, they do not yet provide sufficiently satisfactory performance to compete with the FFT in other complex downstream tasks. Furthermore, the scalability of prompt-based methods~\cite{vpt, vp} is significantly limited, leading to considerable performance degradation as the number of learnable parameters (\ie, prompts) increases, as reported in~\cite{adaptformer}. In contrast, adapter-based approaches~\cite{adaptformer, lora, lorand} incorporate lightweight modules to reduce the number of trainable parameters, maintaining stable performance on a range of trainable parameter scales. However, the adapters always apply the same transformation to input features, ignoring individual characteristics of input instances. This may not be an issue when fully tuning whole networks, but it could be a limiting factor in improving performance in adapter-based PEFT methods. Namely, the inflexibility with respect to input instances exacerbates the transfer capability of adapter-based models with only small learnable parts to downstream tasks, limiting their ability to capture unique and task-specific information. 

Furthermore, the ViT~\cite{vit} used in adapter-based models tends to focus on global information rather than fine local details within an image. While this limitation can be partially addressed by employing the Swin Transformer~\cite{swin}, which utilizes local attention mechanisms, the constraint on the number of learnable parameters in adapter-based approaches still restricts the Swin Transformer's capability to effectively capture local features (Figure~\ref{analysis}). Consequently, this negatively affects the performance in dense prediction tasks that require local details. 

To address the aforementioned issues, we propose a novel PEFT approach, {\bf i}nput-{\bf Con}ditioned trans{\bf Former} ({\bf iConFormer}), which leverages a \emph{dynamic} adapter where parameters are adjusted at the input instance level, unlike existing adapter-based approaches~\cite{adaptformer, lora, lorand}. We introduce an input-Conditioned Network (iCoN) that dynamically generates the parameters for each input feature in the adapter. This approach enables for effectively capturing task-specific and local details for each instance while keeping the number of learnable parameters small. The effectiveness of our method is evidenced by the quantitative analysis in Figure~\ref{analysis_graph}. 
Our method achieves performance competitive to the FFT in both classification and dense prediction tasks including monocular depth estimation, semantic segmentation, and instance segmentation with only additional 1.6\% to 2.8\% backbone parameters. Our method even surpasses FFT for the image classification in CIFAR100~\cite{cifar100} and the challenging instance segmentation task on COCO~\cite{coco}. Additionally, iConFormer also outperforms conventional PEFT methods for monocular depth estimation task on NYU-v2~\cite{nyuv2}, demonstrating the effective utilization of pre-trained backbone parameters with additional learnable parameters dynamically finetuned for specific tasks. We also analyze the capability to capture fine-grained details by visualizing attention maps in Figure~\ref{analysis}. 

In summary, our contributions are threefold:
\begin{itemize}
\item We propose iConFormer to enhance representation learning by dynamically adjusts only a small subset of parameters conditioned on input instances in the PEFT framework.

\item We demonstrate that iConFormer effectively captures fine-grained details with input-Conditioned Network (iCoN), leading to substantial improvements in dense prediction tasks.

\item Through comprehensive experiments on classification, monocular depth estimation, instance segmentation, and semantic segmentation, we show that iConFormer achieves remarkable performance by tuning only 1.6\% to 2.8\% of the Transformer backbone parameters.
\end{itemize}

\begin{figure*}[t]
\centering
\includegraphics[width=0.77\textwidth]{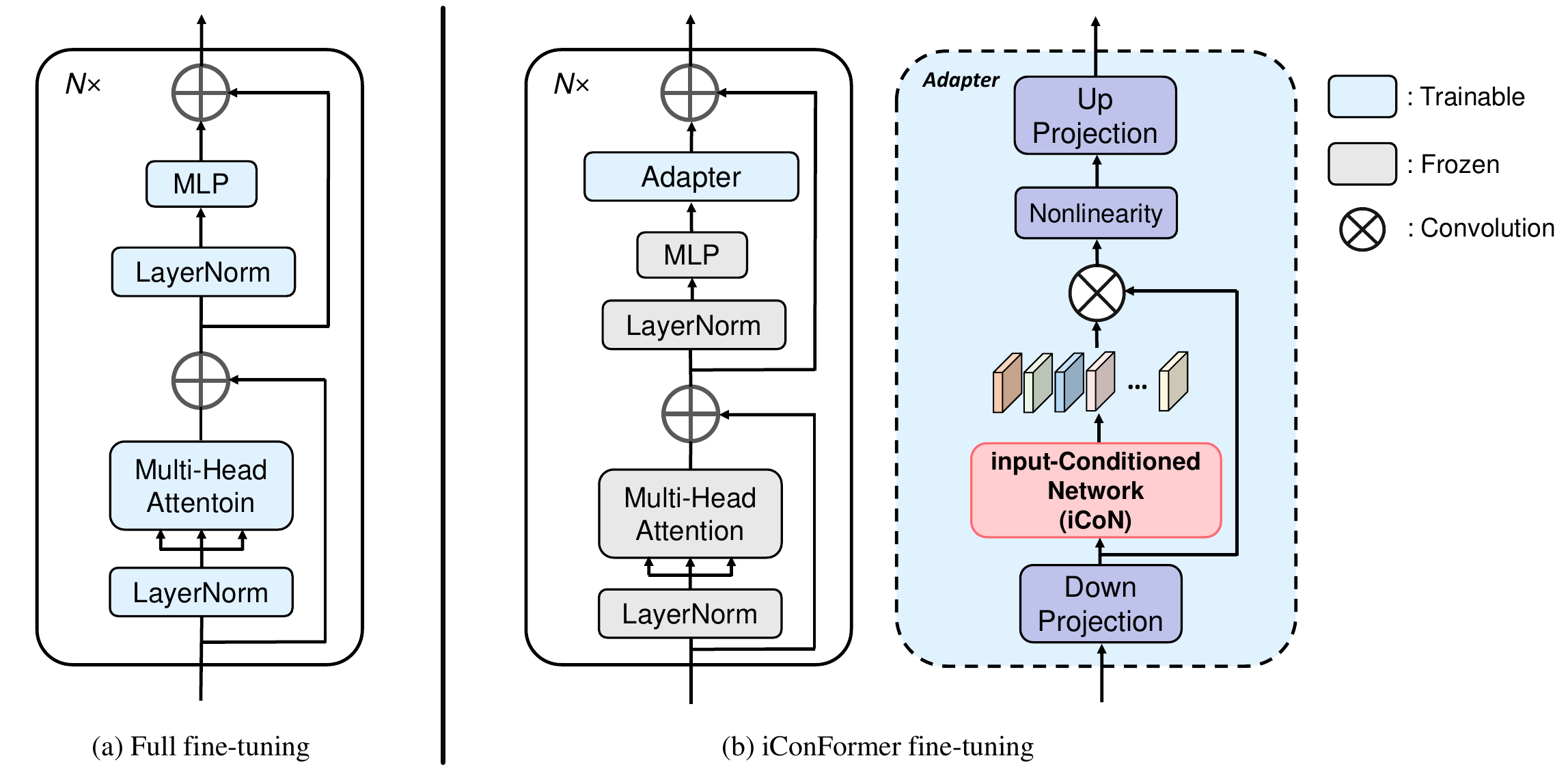} 
\caption{{ \textbf{Comparison of Full Fine-Tuning (FFT) and the proposed Parameter Efficient Fine-Tuning (PEFT) using iConFormer}. (a) FFT, where all parameters are updated during training. (b) Our PEFT (iConFormer), where an \emph{dynamic} adapter is attached sequentially after the MLP layer in the Transformer. Inside the dynamic adapter, an Input-Conditioned Network (iCoN) generates input-conditioned convolutional kernels in a channel basis, which is detailed in Figure~\ref{icon_architecture}. By convolving features with these kernels, iCoN adaptively refines them} in accordance with the specific properties of the input, thereby enhancing the model’s capability to effectively process diverse input data in the downstream tasks.}
\label{architecture}
\vspace{-0.4cm}
\end{figure*}

\section{Related Work}
\label{sec:related}

\subsection{Transformer in Vision}

Transformers, initially designed for Natural Language Processing (NLP) tasks such as machine translation~\cite{transformer} and text generation~\cite{bert}, have achieved significant success in these areas. This success has led to a shift towards computer vision, starting with the Vision Transformer (ViT)~\cite{vit}. Subsequently, various Transformer-based models~\cite{swin, segformer, detr, lee2022knn, lee2023cross, dpt, hong2022cost} have achieved notable advancements in tasks including image classification~\cite{krizhevsky2012imagenet}, semantic segmentation~\cite{long2015fully, chen2017deeplab, cho2024cat, kim2022sequential}, object detection~\cite{girshick2014rich, ren2016faster, redmon2016you}, image restoration~\cite{dong2015image, liu2017unsupervised}, and depth estimation~\cite{godard2019digging, choi2021adaptive}. Furthermore, transformers have significantly advanced vision recognition through large-scale pretraining~\cite{empirical, emerging, mae}. However, their larger size compared to previously prevalent CNN backbones presents challenges for fine-tuning on specific tasks. In this context, our work explores methods to adapt pre-trained transformers into target tasks in a more effective and efficient way.

\subsection{Parameter Efficient Fine Tuning}

Parameter Efficient Fine-Tuning (PEFT) methods enable the adaptation of large pre-trained models~\cite{xie2022simmim, he2022masked, yi2022masked, choi2024emerging, choi2024salience} to specific tasks without the need to train the entire model. In NLP, notable approaches include adapter methods~\cite{adapter_nlp}, which integrate small learnable modules into the model while keeping the pre-trained parameters frozen, with only the added modules being fine-tuned. Additionally, other methods involve tuning specific components such as bias or normalization layers~\cite{bitfit, LN-tuning}, utilizing learnable prompt tokens~\cite{prompt_nlp}, or applying low-rank matrix approximations~\cite{lora} to efficiently update parameters. Recently, a method has been proposed to improve inference time while maintaining the training parameter efficiency by selectively skipping the computation of less important tokens~\cite{coda}.

In computer vision, PEFT techniques inspired by NLP have shown significant progress. VPT~\cite{vpt} is the first method to apply prompt tuning approaches to visual recognition tasks. AdaptFormer~\cite{adaptformer} introduces a parallel adapter framework to enhance the effectiveness of parameter efficient fine-tuning for visual recognition. KAdaptation~\cite{kadaptation} optimizes the adapter using Kronecker products, and SPT~\cite{spt} selectively allocates trainable parameters to important positions under a specified budget. In addition, LoRand~\cite{lorand} employs multi-branch low-rank adapters to achieve impressive performance on dense prediction tasks. Our approach is also based on the adapter framework but introduces input-conditioned kernels for instance-specific adaptation, allowing more precise and flexible fine-tuning with a limited number of learnable parameters.
\section{Preliminary}
\label{sec:prelim}

\subsection{Vision Transformer and its Variants}

Vision Transformer (ViT)~\cite{vit}, modified from the Transformer~\cite{transformer} proposed in NLP, integrates image patches and positional encodings to capture spatial information. It consists of a patch embedding layer and multiple sequential encoder blocks, as depicted in Figure~\ref{architecture} (a).
Given a batch of images $x \in \mathbb{R}^{B\times H\times W\times 3}$, the patch embedding layer transforms $x$ into sequential patches $x_{p}\in \mathbb{R}^{B\times M \times (P^2C)}$, where $H\times W$ is an image resolution, and $B$ is a batch size. $P\times P$ is the resolution of an image patch, $C$ is the output channel, and $M = HW / P^2$ is the number of image patches. The patches are linearly embedded into $D$ dimensions to generate the final input $x_{in}\in \mathbb{R}^{B \times M \times D}$.

In the Transformer encoder block, $x_{in}$ is first normalized using LayerNorm (LN)~\cite{layernorm}, and then processed by a multi-head self-attention layer (MHSA). The output is combined with $x_{in}$ via a residual connection:
\begin{equation}
\acute{x_{in}} = \textrm{Attention}(\textrm{LN}(x_{in}))+x_{in}\,.
\label{eq:mhsa-ln-res}
\end{equation}
\noindent Next, $\acute{x_{in}}$ is normalized and passed through the MLP layer, followed by residual connection:
\begin{align}
&\tilde{x} = \textrm{MLP}(\textrm{LN} (\acute{x_{in}}))\,, \nonumber\\
&x_{out} = \tilde{x} + \acute{x_{in}}\,.
\label{eq:mlp-ln-res}
\end{align}
This process is repeated $N$ times in the encoder block.
In ViT, the self-attention mechanism captures global features by evaluating relationships between all image patches, enabling a comprehensive understanding of complex dependencies. Advancing this approach, Swin Transformer~\cite{swin, swinv2} introduces hierarchical attention with shifted windows, which enhances both computational efficiency and feature representation. Other variants~\cite{pvt, segformer, mask_dino, vit-comer} leverage multi-scale feature extraction for specific vision tasks, improving the adaptability of Transformer models.

\subsection{PEFT Methods}
\begin{figure}[t]
\centering
\includegraphics[width=0.41\textwidth]{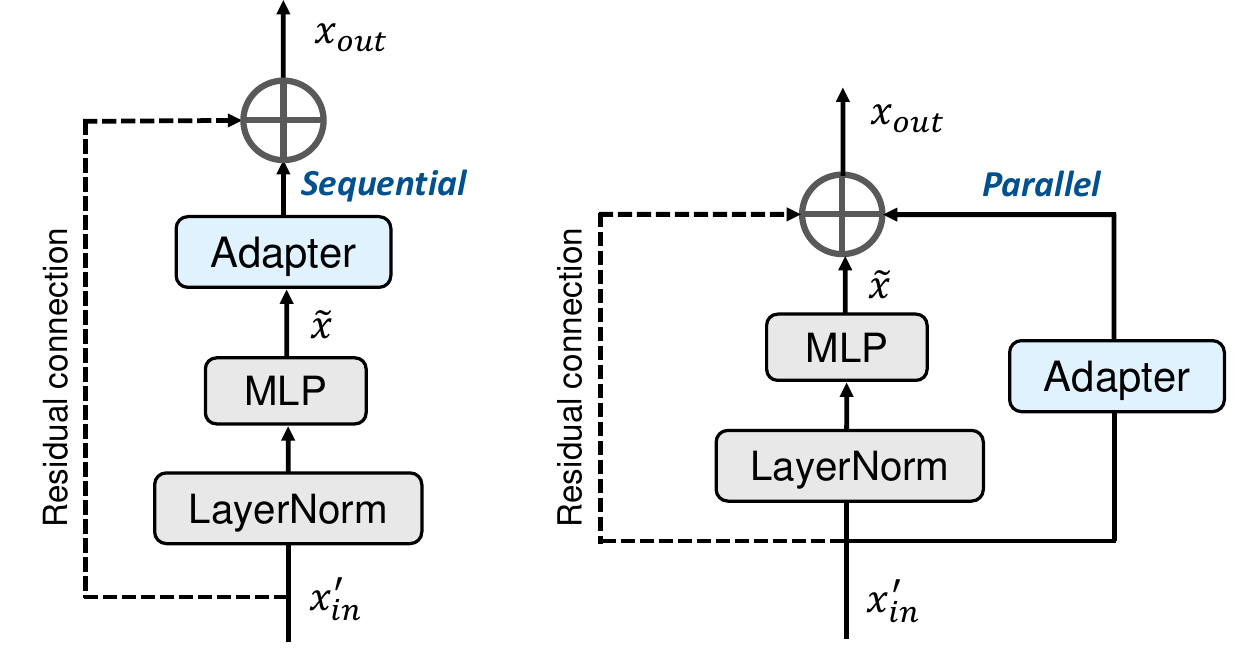} 
\caption{\textbf{Illustration of the sequential and parallel configurations}. The sequential design is shown on the left, and the parallel design on the right.}
\label{Configuration}
\vspace{-0.4cm}
\end{figure}

Parameter efficient fine-tuning (PEFT) methods, such as prompt tuning~\cite{prompt_nlp, vpt}, low-rank adaptation~\cite{lora}, and adapters~\cite{adapter_nlp, adaptformer}, are designed to reduce the number of trainable parameters needed for fine-tuning large models.  
Here, we briefly review adapter-based PEFT methods related to our approach, which will be detailed in the following section. 
The adapter introduces small, trainable modules between the layers of the pre-trained model. It can be integrated in sequential or parallel configurations as shown in Figure~\ref{Configuration}. For instance, if the original architecture processes the features as described in \eqref{eq:mlp-ln-res}, conventional adapters modify this transformation to 
\begin{equation}
x_{out} =
\begin{cases}
\gamma \cdot \text{Up}(\sigma (\text{Down}(\tilde{x}))) + \acute{x_{in}}, &  \text{(Sequential)} \\
\tilde{x} + \gamma \cdot \text{Up}(\sigma (\text{Down}(\acute{x_{in}}))) + \acute{x_{in}}, & \quad \text{(Parallel)} \\
\end{cases}
\label{eq:conventional_adapter}
\end{equation}
where $\text{Down}(\cdot)$ represents the down-projection of the input features, $\text{Up}(\cdot)$ indicates the up-projection back to the original space, and $\sigma$ denotes an activation function. Here, $\gamma$ is a weighting factor that adjusts the contribution of the adapter output. These approaches allow for task adaptation while minimizing the number of learnable parameters that need to be updated during model training.

\section{Proposed Method}
\label{sec:method}

\subsection{Motivation and Overview}
Conventional adapter-based methods~\cite{adapter_nlp, adaptformer} rely on \emph{static} parameter-tuning, where the learnable modules such as `$\text{Up}$' and `$\text{Down}$' added to the original transformation are always static with respect to the input feature $\tilde{x}$ or $\acute{x_{in}}$, as described in \eqref{eq:conventional_adapter}. Thus, the same transformation is applied to all instances regardless of instance-specific characteristics, limiting the ability to adapt to input feature distributions in the constrained training environment where only a few number of parameters are tuned for various downstream tasks. Namely, in the PEFT that allows for updating only a minimal set of parameters, the static transformations become insufficient for handling diverse input variations, necessitating an instance-level mechanism that \emph{dynamically} adjusts the learnable parameters conditioned on the input features.

To this end, we incorporate dynamic kernel generation, enabling instance-aware transformations while maintaining parameter efficiency under PEFT constraints. Specifically, we introduce iConFormer, a novel framework incorporating the input-Conditioned Network (iCoN), as illustrated in Figure~\ref{architecture} (b). Unlike conventional static adapters, iConFormer dynamically generates input-conditioned convolution kernels tailored to the unique characteristics of each instance, enabling more precise and flexible adaptation. By introducing dynamic kernel generation, the proposed method enhances feature extraction while incorporating locality inductive biases into the pretrained Transformer encoder, allowing the model to dynamically capture both global and local features.

\begin{figure}[t]
\centering
\includegraphics[width=1.0\columnwidth,height=0.37 \columnwidth]{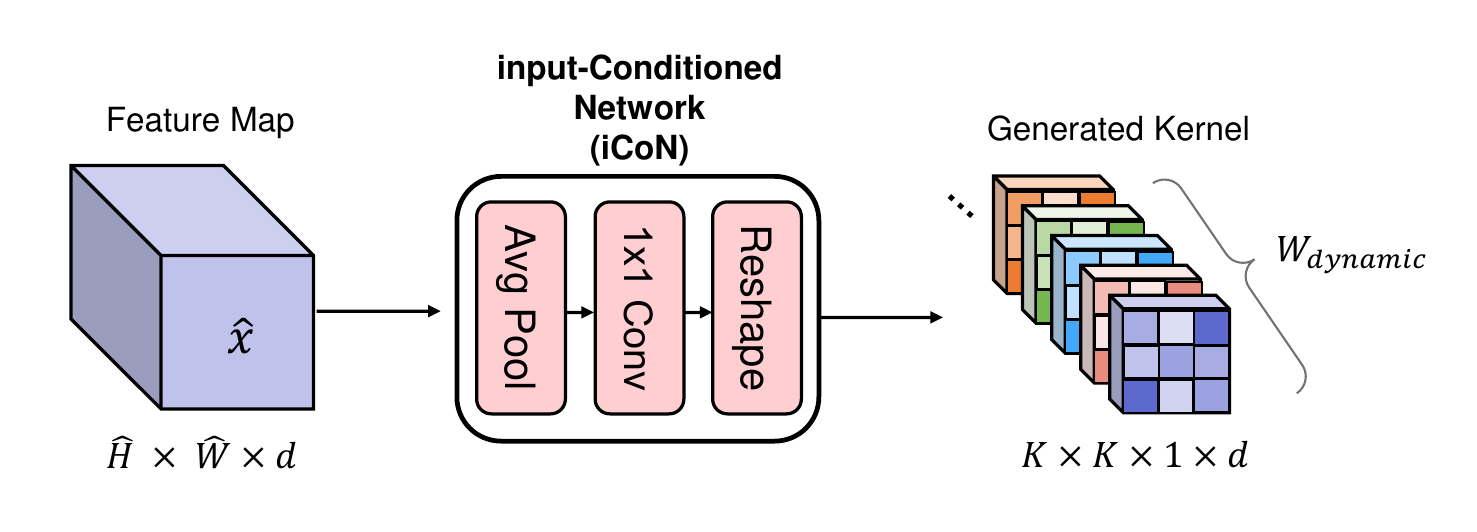}
\caption{\textbf{Architecture of input-Conditioned Network (iCoN)}. The down-projected feature map $\hat{x}$ is used to dynamically generate channel-wise convolution kernels through the iCoN.}
\label{icon_architecture}
\vspace{-0.4cm}
\end{figure}

\subsection{Input-Conditioned Network (iCoN)}

The iCoN is a key component of the iConFormer framework. Inspired by the concept of dynamic filter networks~\cite{dfn}, iCoN employs a channel-wise mechanism to dynamically generate convolutional kernels that are tailored to the unique characteristics of each input. This approach enhances parameter efficiency while maintaining the flexibility needed to capture diverse features effectively. Formally, the iCoN module generates channel-wise convolutional kernels using input features from the MLP output, denoted as $\tilde{x}\in \mathbb{R}^{B \times M \times D}$ of \eqref{eq:mlp-ln-res}. First, the feature $\tilde{x}$ is down-projected to $d$ channels and reshaped from $M$ into the spatial dimensions $\hat{H} \times \hat{W}$, which correspond to $H/P \times W/P$. This reshaping rearranges the patches into a spatial grid, producing the feature $\hat{x}\in\mathbb{R}^{B \times \hat{H} \times \hat{W} \times d}$. Subsequently, the iCoN module generates dynamic convolutional kernels from the reshaped feature $\hat{x}$. This process is mathematically represented as:
\begin{equation}
\hat{x} = \text{Reshape}(\text{Down}(\tilde{x})),
\end{equation}
\begin{equation}
W_{dynamic} = \text{iCoN}(\hat{x}),
\end{equation}
where $W_{dynamic}\in \mathbb{R}^{B \times d \times K \times K}$ is the dynamically generated kernel and $K$ is the kernel size.  For our implementation, we set $K$ to 3 and $d$ to 64 ($d \ll D$). 

Figure~\ref{icon_architecture} illustrates the process of dynamically generating convolutional kernels in the iCoN module. It first applies spatial average pooling to extract global contextual information from $\hat{x}$. The pooled features are then passed through a lightweight transformation, which learns a mapping function to parameterize the convolutional kernel weights. Finally, the transformed features are reshaped into the channel-wise convolutional kernel $W_{dynamic}$. The convolution kernel generated conditioned on the input feature $\hat{x}$ is then applied to $\hat{x}$ through a channel-wise convolution operation.
Afterward, the feature is reshaped back to $\mathbb{R}^{B \times M \times d }$, followed by applying a non-linear activation function and up-projection to restore the channel dimension to the original size $D$. This process produces the final output of the adapter as follows:
\begin{equation}
x_{A} = \text{Up}( \sigma (\text{Reshape}(\hat{x} \otimes W_{dynamic}))),
\label{eq:input_conv}
\end{equation}
where ${\otimes}$ denotes the channel-wise convolution operation and ${\sigma}$ represents the GeLU activation function. 
The adapter employs a sequential structure as illustrated in Figure~\ref{Configuration} (left) to effectively integrate with the model’s feature processing, and a residual connection is applied to enhance model robustness:
\begin{equation}
x_{out} = \gamma \cdot x_{A} + \acute{x_{in}} \,,
\end{equation}

\noindent where $\gamma$ is a weight that adjusts the impact of the adapter features. This weight is a learnable scalar, optimized during training.

By dynamically generating convolutional kernels, iCoN allows the model to effectively adapt to the input structure while enriching the feature representation with both global context and local details. Here, it should be noted that the standard convolution operation can also be applied to the Transformer encoder to inject the locality, but the dynamic adaptation to the input instance is more crucial to the PEFT framework for various downstream tasks. In contrast to standard convolutional kernels, iCoN dynamically adjusts its kernels based on the spatial structure of each input, resulting in more precise and context-aware feature extraction. This mechanism allows iCoN to capture high-frequency variations and subtle patterns that the standard convolution layers struggle to represent effectively. This is also validated in experiments (Table~\ref{Ablation3}).

It is worthy of noting that while our approach generates dynamic kernels, it remains computationally comparable to conventional adapter-based PEFT methods~\cite{adaptformer, adapter_nlp}. This efficiency is primarily attributed to two factors. First, while iCoN generates dynamic kernels, the network responsible for the kernel generation retains static learnable parameters, ensuring that only the output kernels adapt to instance without significantly increasing parameter overhead. Second, to further reduce computational complexity while preserving adaptive capacity, iCoN employs the channel-wise convolutions instead of the original convolutions, as shown in Figure~\ref{icon_architecture}.

\begin{figure}[t]
\centering
\includegraphics[width=0.9\columnwidth]{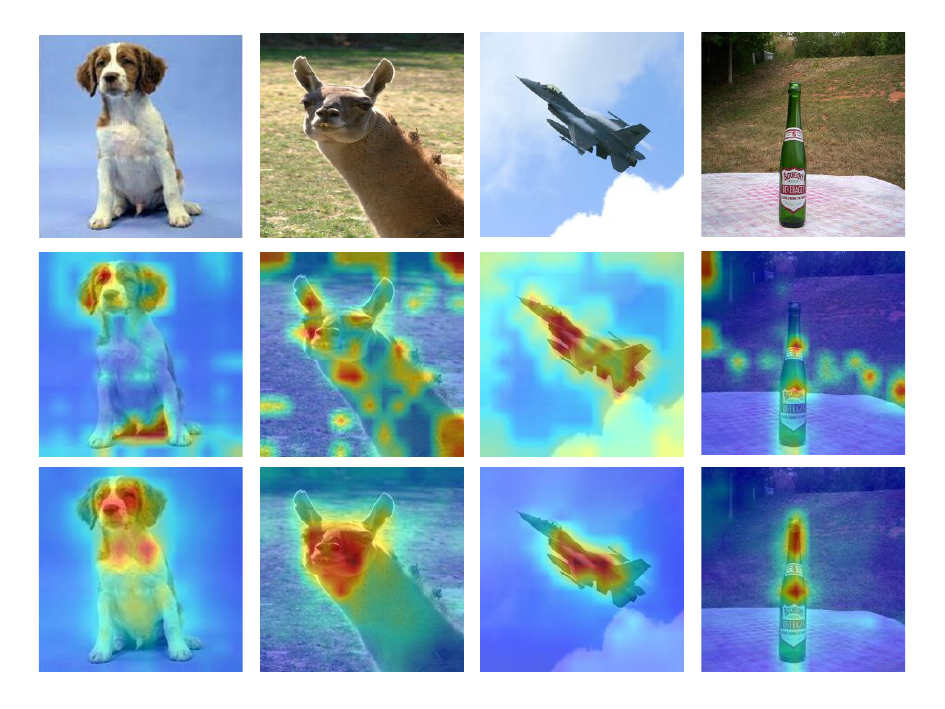}
\vspace{-0.4cm}
\caption{\textbf{Comparison of Attention Maps from AdaptFormer and iConFormer}. We visualize the attention maps using attention rollout~\cite{rollout}. The top row represents input images, and the middle and bottom rows present the attention maps generated by AdaptFormer~\cite{adaptformer} and iConFormer, respectively, with both using Swin Transformer backbone~\cite{swin}. iConFormer more accurately delineates object regions and captures fine-grained semantics, compared to the AdaptFormer.}
\label{analysis}
\vspace{-0.3cm}
\end{figure}


\subsection {Visual Analysis of Local Representation}
To evaluate the effectiveness of capturing both local and global information, Figure~\ref{analysis} visualizes the attention maps that provide insight where the model focuses. The attention maps are generated using attention rollout~\cite{rollout} with AdaptFormer~\cite{adaptformer} and iConFormer, employing the Swin Transformer~\cite{swin} as the backbone. Attention rollout computes token attentions by recursively multiplying attention matrices across layers, revealing how attention is distributed across different regions of an input image. AdaptFormer adopts a standard pipeline consisting of down-projection, non-linear activation, and up-projection based on static weight parameters that are not dynamically adjusted conditioned on input features. While the attention maps of AdaptFormer exhibit ambiguous and scattered attention distributions with limitations in precisely capturing local features such as object edges, the attention maps of iConFormer are significantly more focused and better aligned with objects.

The iCoN used in the iConFormer dynamically generates convolutional kernels tailored to the input features, enabling the iConFormer to capture detailed spatial features while preserving overall contextual awareness. By focusing on these salient details, the proposed method demonstrates significant improvements in processing complex input data, leading to enhanced accuracy and robustness in dense prediction tasks.
\section{Experiments}
\label{sec:experiments}

\begin{table}[t]
\caption{\textbf{Performance evaluation of image classification.} 
We report the absolute Top-1 accuracy on the CIFAR-100, SVHN, and Food-101 datasets. $^\dagger$ indicates a learning rate reduced to $0.1\times$ due to unstable training. Additionally, FLOPs are measured with a batch size of 2.}
\label{classification}
\centering
\scriptsize\addtolength{\tabcolsep}{-3pt}

\begin{tabular}{l|c|c|c c c}
\toprule                                        
Method                        & Params (M)&FLOPs (G)  & CIFAR-100 & SVHN & Food-101   \\
\midrule
\rowcolor{gray!20}
Full      & 86.04 (100\%)&   35.16  & 85.90   & 97.67$^{\dagger}$ & 90.09$^{\dagger}$  \\         
Linear   & 0.07 (0.08\%)& 35.16& 69.83 & 66.91 & 69.74 \\ 
VPT~\cite{vpt} & 0.08 (0.09\%)&35.35 & 82.44 & 94.02 & 82.98 \\ 
AdaptFormer~\cite{adaptformer} & 1.26 (1.46\%)&35.63 & 85.90 & 96.89 & 87.61 \\
Adapter~\cite{adapter_nlp}&  2.46 (2.86\%)&36.13 & 86.65 &  97.09&86.89  \\
\midrule
\textbf{iConFormer} & 1.71 (1.98\%)& 35.65& \textbf{86.94} &  \textbf{97.38} & \textbf{87.97}\\

\bottomrule
\end{tabular}
\vspace{-0.3cm}
\end{table}

\subsection{Experimental Settings}

\begin{table*}
\caption{\textbf{Performance evaluation of monocular depth estimation on the NYU-v2 dataset.} The results show comparisons of iConFormer with various parameter-efficient fine-tuning approaches. Results with the symbol $\uparrow$ / $\downarrow$ indicate higher/lower is better.}
\label{depth:nyuv2}
\centering
\scriptsize\addtolength{\tabcolsep}{5pt}
\begin{tabular}{l|c|c c c c c c}
\toprule                                        
Method                        & Params (M) & $\delta_{1}\uparrow$ & $\delta_{2}\uparrow$  &$\delta_{3}\uparrow$ & AbsRel$\downarrow$  &RMSE$\downarrow$   & $\textit{log}_{10}\downarrow$     \\
\midrule
\rowcolor{gray!20}
Full      & 86.9 (100\%)     & 0.935   & 0.991 & 0.998 & 0.044 & 0.304 & 0.109 \\         
Fixed   & 0 (0\%) & 0.454 &0.748 & 0.896 & 0.374 & 1.018 &0.382   \\ 
Partial-l~\cite{partial} & 12.62 (14.52\%) & 0.492 & 0.797 & 0.928 &0.307 &0.925 &0.346 \\
BitFit~\cite{bitfit} & 0.14 (0.16\%) & 0.823 & 0.969 & 0.992 & 0.144 & 0.466 & 0.169 \\
LN-tuning~\cite{LN-tuning} & 0.05 (0.06\%) & 0.802 &0.963 &0.990 &0.152 & 0.496 &0.180 \\
Adapter~\cite{adapter_nlp} & 3.11 (3.45\%) & 0.901 & 0.987 & 0.997 & 0.104 &0.361 &0.130 \\
AdaptFormer~\cite{adaptformer}& 1.55 (1.76\%) & 0.845 & 0.975 & 0.994 & 0.134 & 0.432 &0.159 \\
LoRA~\cite{lora} & 3.08 (3.42\%) & 0.439 & 0.733 & 0.885 & 0.402 & 1.045 & 0.395 \\

\midrule
\textbf{iConFormer} & 2.48 (2.68\%) & \textbf{0.914} & \textbf{0.988} & \textbf{0.998} & \textbf{0.098} & \textbf{0.342} & \textbf{0.122}\\

\bottomrule
\end{tabular}
\vspace{-0.3cm}
\end{table*}

\begin{table}
\caption{\textbf{Performance evaluation of semantic segmentation on the ADE20K dataset.} The results show comparisons of iConFormer with various adapter-based approaches.}
\label{semantic:ADE20k}
\centering
\scriptsize\addtolength{\tabcolsep}{5pt}
\begin{tabular}{l|c|c}
\toprule                                        
Method                        & Params (M) & $\text{mIoU}\uparrow$  \\
\midrule
\rowcolor{gray!20}
Full      & 198.58 (100\%)     & 51.50 \%   \\         
Fixed   &  0 (0\%) & 32.21 \%  \\ 
Adapter-B~\cite{adapter_nlp} & 32.04 (16.13\%) & 46.23 \%  \\
Adapter-T~\cite{adapter_nlp} & 16.04 (8.08\%) & 43.51 \%  \\
AdaptFormer~\cite{adaptformer} & 2.34 (1.18\%) & 49.85 \%  \\
LoRA~\cite{lora} & 4.57 (2.31\%) & 49.48 \% \\
LoRand~\cite{lorand} & 3.59 (1.84\%) & 50.67 \%  \\

\midrule
\textbf{iConFormer} & 3.26 (1.65\%) & \textbf{51.17\%} \\
\bottomrule
\end{tabular}
\vspace{-0.1cm}
\end{table}

\begin{table}
\caption{\textbf{Performance evaluation of instance segmentation on the COCO dataset.} The results show comparisons of iConFormer with various adapter-based approaches.}
\label{instance:coco}
\centering
\scriptsize\addtolength{\tabcolsep}{5pt}
\begin{tabular}{l|c|c c}
\toprule                                        
Method   & Params (M) & $\textrm{AP}_{Box}\uparrow$ & $\textrm{AP}_{Mask}\uparrow$  \\
\midrule
\rowcolor{gray!20}
Full      & 89.14 (100\%)     & 51.90 \%  & 45.00 \%\\     
Fixed   &  0 (0\%) & 15.30 \%  & 10.80 \%\\ 
Adapter-B~\cite{adapter_nlp} & 14.38 (16.13\%) & 46.50 \%  &  40.20 \%\\
Adapter-T~\cite{adapter_nlp} & 7.20 (8.08\%) & 43.20 \% & 38.70 \% \\
AdaptFormer~\cite{adaptformer} & 1.60 (1.79\%) & 48.90 \%  & 42.50 \%\\
LoRA~\cite{lora} & 3.08 (3.43\%) & 47.50 \% & 41.50 \%\\
LoRand~\cite{lorand} & 2.39 (2.76\%) & 51.10 \%  & 44.10 \%\\

\midrule
\textbf{iConFormer} & 2.48 (2.78\%) & \textbf{52.90\%} & \textbf{45.90\%}\\

\bottomrule
\end{tabular}
\vspace{-0.4cm}
\end{table}

\paragraph{Datasets and Downstream Tasks} To evaluate the performance of iConFormer, we conducted comprehensive experiments on both image classification and dense prediction tasks, including monocular depth estimation, semantic segmentation, and instance segmentation. Implementation details can be found in the supplementary material. The datasets used in the experiments are as follows:
\begin{itemize}
\item \textbf{Image Classification}: CIFAR-100 dataset~\cite{cifar100} consists of 50,000 training images and 10,000 validation images, each with a resolution of 32×32, categorized into 100 classes. The SVHN dataset~\cite{svhn} includes over 600,000 labeled images for digit classification, comprising 73,257 training samples, 26,032 test samples, and 531,131 additional training images. The Food-101 dataset~\cite{food101} contains 101,000 images across 101 food categories, with each category having 750 training samples and 250 test samples.
\item \textbf{Monocular Depth Estimation}:  NYU-v2~\cite{nyuv2} with diverse indoor scenes and KITTI~\cite{kitti} with high-resolution outdoor driving scenes are benchmark datasets for depth estimation.
For experiments, we used the standard splits and evaluate using Root Mean Square Error (RMSE). NYU-v2 images were cropped to 352 × 352 pixels, while KITTI images were cropped to 480 × 480 pixels.
\item \textbf{Semantic Segmentation}: ADE20K~\cite{ade20k} is a widely used semantic segmentation dataset with 20,000 training and 2,000 validation images. For our experiments, we utilized UperNet~\cite{upernet} as the framework and evaluated performance using the mean Intersection over Union (mIoU) metric.
\item \textbf{Instance Segmentation}: MS COCO~\cite{coco} is a prominent dataset for instance segmentation, with 118,000 training and 5,000 validation images. We used Cascade Mask R-CNN~\cite{cascade, maskrcnn} as a task-specific decoder and measured performance with Average Precision for bounding boxes ($\textrm{AP}_{Box}$) and masks ($\textrm{AP}_{Mask}$).
\end{itemize}

\vspace{-0.3cm}

\paragraph{Pretrained Backbones} For a fair comparison with FFT baseline and current PEFT methods, we used different pre-trained backbones depending on the tasks. In the semantic segmentation and instance segmentation tasks, Swin Transformer backbones~\cite{swin}, pre-trained on ImageNet-22k dataset~\cite{imagenet}, were used~\cite{openmmlab}. Specifically, we used the Swin-Large backbone for semantic segmentation and the Swin-Base backbone for instance segmentation. For the monocular depth estimation, we utilized the standard Swin-V2-Base backbone~\cite{swinv2} pre-trained using the MIM~\cite{mim}. For the classification task, we adopted the ViT backbone~\cite{vit} pre-trained using MAE~\cite{mae}.

\vspace{-0.3cm}

\paragraph{Baseline Methods} For the image classification task, we used the same set of comparison models as~\cite{adaptformer}, and additionally included the Adapter method from~\cite{adapter_nlp}. In the monocular depth estimation, we included comparisons with partial tuning methods such as BiTFiT~\cite{bitfit}, LN-Tuning~\cite{LN-tuning}. We also evaluated against Partial-l~\cite{partial}, which fine-tunes only the final block of the backbone and parameters outside the backbone. For comparison with adapter-based methods, we included recent approaches such as Adapter~\cite{adapter_nlp}, AdaptFormer~\cite{adaptformer}, LoRA~\cite{lora}, and LoRand~\cite{lorand}.
In the semantic segmentation and instance segmentation tasks, we configured the Adapter~\cite{adapter_nlp} following~\cite{lorand} for a fair comparison, setting the intermediate layer dimension to half of the input dimension for `Adapter-B' and to a quarter for `Adapter-T'. Additionally, we included `Fixed' in all dense prediction tasks, freezing the pre-trained Transformer encoder while training other parts of the architecture (\ie, task decoder). Across all tasks, we also included `Full', which indicates full fine-tuning (FFT) as an upper bound on performance.

\subsection{Main Results}

\paragraph{Image Classification}
We evaluated various fine-tuning approaches using ViT backbone~\cite{vit} pre-trained via self-supervised learning paradigms~\cite{grill2020bootstrap, chen2021exploring, ema1, ema2}, as detailed in Table~\ref{classification}. The results demonstrate that the iConFormer consistently outperforms linear probing, Visual Prompt Tuning (VPT) methods, and recently proposed adapter-based techniques. Specifically, the iConFormer achieves performance improvements of 4.5\%, 3.36\%, and 4.99\% over VPT on the image benchmarks CIFAR-100, SVHN, and Food-101, respectively. Furthermore, when compared to recent adapter-based methods such as Adapter~\cite{adapter_nlp} and AdaptFormer~\cite{adaptformer}, iConFormer shows up to 1.04\%, 0.49\%, and 1.08\% higher accuracy, respectively. Notably, iConFormer also surpasses the FFT approach by more than 1\% Top-1 accuracy on the CIFAR-100 dataset. Additionally, while consistently delivering better performance, iConFormer demonstrates computational efficiency with 35.65 GFLOPs, which is comparable to AdaptFormer (35.63G) and lower than Adapter (36.13G). In summary, our approach outperforms recent adapter-based approaches with similar computational efficiency, and provides comparable performance to the FFT despite tuning only 2\% of the parameters used in the FFT.

\vspace{-0.3cm}

\paragraph{Monocular Depth Estimation}
Table~\ref{depth:nyuv2} presents the performance results for the NYU-v2 datasets. As shown in the tables, the iConFormer outperforms other PEFT methods in all metrics, with the RMSE value being within 0.04 of the FFT performance. Moreover, the iConFormer shows an RMSE improvement of up to 0.2 compared to partial tuning methods, and an enhancement of up to 0.3 RMSE compared to adapter-based methods such as Adapter~\cite{adapter_nlp}, AdaptFormer~\cite{adaptformer}, and LoRA~\cite{lora}. These results suggest that iConFormer’s capability to generate and apply input-conditioned kernels significantly contributes to the performance in the monocular depth estimation task. Additional results on the KITTI dataset are presented in the supplemental material.

\vspace{-0.3cm}

\paragraph{Semantic Segmentation}
We present the results of the semantic segmentation task on the ADE20K dataset~\cite{ade20k} in Table~\ref{semantic:ADE20k}. By fine-tuning fewer than 3.3 million backbone parameters, the proposed method achieves 51.17\% mIoU on ADE20K, which is about 0.3\% lower than the FFT. Moreover, the iConFormer requires fewer learnable parameters compared to most adapter-based methods while still achieving superior performance. These results suggest that iConFormer effectively utilizes a limited subset of parameters to capture task-specific information and learn detailed features.

\vspace{-0.3cm}

\paragraph{Instance Segmentation}
Table~\ref{instance:coco} presents the instance segmentation results on the COCO dataset. Our method demonstrates significant performance gains by training only 2.78\% of the total backbone parameters, surpassing both existing adapter-based methods and FFT. Specifically, it achieves 1.0\% improvement in $\text{AP}_{box}$ and 0.9\% improvement in $\text{AP}_{mask}$ compared to the FFT. These results reveal the advantages of the proposed method and demonstrate its superiority over the FFT in terms of both storage efficiency and performance. Additionally, these findings suggest that iConFormer optimizes resource utilization through its dynamic kernel approach.

\begin{table}
\caption{\textbf{Ablation study of the sequential and parallel configurations on dense prediction tasks.} Results are presented in the order of NYU-v2, ADE20K, and COCO, from left to right.}
\label{Ablation2}
\centering
\scriptsize\addtolength{\tabcolsep}{3pt}

\begin{tabular}{c|c|c|cc}
\toprule
Configuration 
              & RMSE ↓                        & mIoU ↑                       & $\textrm{AP}_{Box} \uparrow$ & $\textrm{AP}_{Mask} \uparrow$ \\ 
\midrule
Parallel      & 0.357                         & 50.85 \%                     & 51.20 \%           & 44.60 \%             \\
\rowcolor{gray!20}
Sequential    & 0.342                         & 51.17 \%                     & 52.90 \%           & 45.90 \%             \\
\bottomrule
\end{tabular}
\vspace{-0.5cm}
\end{table}

\begin{table*}[t]
\caption{\textbf{Ablation study of input-Conditioned kernel size in iCoN.} We perform a quantitative comparison of different kernel sizes across dense prediction datasets. Results with the symbol $\uparrow$ / $\downarrow$ indicate the higher/lower is the better.}
\label{Ablation1}
\centering
\scriptsize\addtolength{\tabcolsep}{5pt}

\centering
\begin{tabular}{c|cc|cc|ccc}
\toprule
\multirow{2}{*}{Kernel Size} & 
\multicolumn{2}{c|}{{KITTI}} & 
\multicolumn{2}{c|}{{ADE20K}} & 
\multicolumn{3}{c}{{COCO}} \\ \cline{2-8}
& {Params (M)} & {RMSE ↓} &{Params (M)} &{mIoU ↑} & {Params (M)} & {$\textrm{AP}_{Box} \uparrow$} & {$\textrm{AP}_{Mask} \uparrow$} \\ 
\midrule
\rowcolor{gray!20}
3 $\times$ 3 & 2.48 (2.68\%) & 2.302 & 3.26 (1.65\%) & 51.17 \% & 2.48 (2.78\%) & 52.90 \% & 45.90 \% \\
5 $\times$ 5 & 4.07 (4.48\%) & 2.314 & 4.86 (2.43\%) & 50.95 \% & 4.07 (4.49\%) & 52.60 \% & 45.70 \% \\
7 $\times$ 7 & 6.47 (6.93\%) & 2.320 & 7.26 (3.59\%) & 50.87 \% & 6.47 (6.94\%) & 52.60 \% & 45.70 \% \\
\bottomrule
\end{tabular}
\vspace{-0.3cm}
\end{table*}

\begin{table}
\caption{\textbf{Ablation study on the effect of the Input-Conditioned Kernel in dense prediction tasks.}
Results are presented in the order of NYU-v2, ADE20K, and COCO datasets, from top to bottom. Both kernel types uses a 3$\times$3 kernel size for all experiments.}
\label{Ablation3}

\centering
\scriptsize\addtolength{\tabcolsep}{10pt}

\begin{tabular}{c|c|c }
\toprule                                        
Kernel Type                   & Params (M) & RMSE$\downarrow$    \\
\midrule
Standard Conv &  2.44 (2.64\%)& 1.029\\    
\rowcolor{gray!20}
input-Conditioned Conv  & 2.48 (2.68\%)  & 0.342   \\ 
\bottomrule
\end{tabular}

\vspace{0.15cm}

\centering
\begin{tabular}{c|c|c}
\toprule                                        
Kernel Type                      & Params (M) & $\text{mIoU}\uparrow$     \\
\midrule
Standard Conv   & 3.25 (1.64\%) & 50.02 \%\\    
\rowcolor{gray!20}
input-Conditioned Conv  & 3.26 (1.65\%) & 51.17 \%   \\ 
\bottomrule
\end{tabular} \\
\vspace{0.15cm}
\centering
\scriptsize\addtolength{\tabcolsep}{-9pt}
\begin{tabular}{c|c|c|c}
\toprule                                        
Kernel Type                      & Params (M) & $\textrm{AP}_{Box}\uparrow$ & $\textrm{AP}_{Mask}\uparrow$     \\
\midrule
Standard Conv   & 2.46 (2.76\%) &50.20 \%  &43.50 \%\\    
\rowcolor{gray!20}
input-Conditioned Conv  &  2.48 (2.78\%) &52.90 \% & 45.90 \%  \\ 
\bottomrule
\end{tabular}
\vspace{-0.3cm}
\end{table}

\subsection{Ablation Studies}

We conducted ablation studies to explore various aspects of the iConFormer and identify key factors that contribute to its performance. All ablation experiments were conducted using the dense predictions tasks.

\vspace{-0.3cm}

\paragraph{iConFormer Configuration}
We investigated the performance by comparing sequential and parallel configurations, as illustrated in Figure~\ref{Configuration}, where the distinction is based on the placement of the Adapter within the Transformer block. As demonstrated in Table~\ref{Ablation2}, the sequential form significantly outperforms the parallel form for all dense tasks. The reason might be: 
(1) the sequential design processes each layer's output in a progressive manner, facilitating deeper feature representations and gradual refinement of complex patterns; (2) the parallel design processes outputs simultaneously, which results in limited inter-layer interaction, weakening the information flow and hindering the model’s capacity to capture intricate features. Therefore, we adopted the sequential design as the default configuration for iConFormer, given its demonstrated superior performance.

\vspace{-0.3cm}

\paragraph{input-Conditioned Kernel Size}
In Table~\ref{Ablation1}, we present an ablation study on the size of the input-conditioned convolution on dense prediction tasks. 
Experiments with kernel sizes of 3$\times$3, 5$\times$5, and 7$\times$7 demonstrate that the 3×3 kernel consistently achieves competitive results with a relatively small number of parameters.
Notably, on the KITTI dataset, the RMSE slightly improves as the kernel size decreases, with 3$\times$3 kernel achieving the lowest RMSE. Similarly, on the ADE20K and COCO datasets, the 3$\times$3 kernel consistently outperforms the 5$\times$5 and 7$\times$7 variants in both mIoU and AP. These results indicate that 3$\times$3 kernel captures essential local features effectively while maintaining computational efficiency. Given that the performance across kernel sizes is quite similar, 3$\times$3 kernel was adopted for its efficiency, providing a balanced trade-off between accuracy and computational cost for the input-conditioned kernels of iCoN.

\vspace{-0.3cm}

\paragraph{Effect of input-Conditioned Kernel}
We investigated the effect of using the input-conditioned convolution $W_{dynamic}$ of \eqref{eq:input_conv} in the iCoN for dense prediction tasks. In Table~\ref{Ablation3}, we compared the performance when using the standard convolution and the input-conditioned convolution for the NYU-v2 dataset (monocular depth estimation), the ADE20K dataset (semantic segmentation), and the COCO dataset (instance segmentation). For a fair comparison, we set to $3\times3$ kernel for both cases, ensuring the same local receptive field. The input-conditioned kernel consistently outperforms the standard variant for all tasks, improving mIoU by about 1.2\% on ADE20K and AP by about 2.5\% on COCO, and reducing RMSE by about 0.7 on NYU-v2.

To further analyze the performance of the standard convolution, we extended the comparison to existing PEFT methods. 
In Table~\ref{depth:nyuv2}, most existing PEFT methods achieve a lower RMSE than the standard convolution. In Table~\ref{semantic:ADE20k} and \ref{instance:coco}, compared to LoRand~\cite{lorand}, the standard convolution achieves approximately 0.6\% lower mIoU on ADE20K and about 1\% lower AP on COCO, respectively. This indicates that simply applying the standard convolution to our framework is not so effective and the adaptive nature of the input-conditioned convolution, where kernel weights are dynamically modulated for input features, is more crucial to capture local details in dense prediction tasks.

\section{Conclusion}
\label{sec:conclusion}

In this work, we have presented iConFormer that leverages a parameter-efficient input-conditioned adapter to effectively capture task-specific features and local information with a limited number of learnable parameters in fine-tuning the models for various downstream tasks. iConFormer demonstrates performance comparable to full fine-tuning across image classification, monocular depth estimation, semantic segmentation, and instance segmentation tasks, by tuning only 1.6\% to 2.8\% of the backbone parameters. iConFormer effectively addresses the limitations of conventional adapter methods and provides superior performances in all tasks. Although our current focus is on vision recognition tasks, we plan to extend iConFormer to other domains such as natural language processing and multi-modal tasks in future work. We anticipate that this extension will inspire further research into efficient adaptation methods and contribute to developing robust solutions across a variety of applications.

{
    \small
    \bibliographystyle{ieeenat_fullname}
    \bibliography{main}

\begin{thebibliography}{68}
\providecommand{\natexlab}[1]{#1}
\providecommand{\url}[1]{\texttt{#1}}
\expandafter\ifx\csname urlstyle\endcsname\relax
  \providecommand{\doi}[1]{doi: #1}\else
  \providecommand{\doi}{doi: \begingroup \urlstyle{rm}\Url}\fi

\bibitem[Abnar and Zuidema(2020)]{rollout}
Samira Abnar and Willem Zuidema.
\newblock Quantifying attention flow in transformers, 2020.

\bibitem[Ba et~al.(2016)Ba, Kiros, and Hinton]{layernorm}
Jimmy~Lei Ba, Jamie~Ryan Kiros, and Geoffrey~E Hinton.
\newblock Layer normalization.
\newblock \emph{arXiv preprint arXiv:1607.06450}, 2016.

\bibitem[Bahng et~al.(2022)Bahng, Jahanian, Sankaranarayanan, and Isola]{vp}
Hyojin Bahng, Ali Jahanian, Swami Sankaranarayanan, and Phillip Isola.
\newblock Exploring visual prompts for adapting large-scale models.
\newblock \emph{arXiv preprint arXiv:2203.17274}, 2022.

\bibitem[Bossard et~al.(2014)Bossard, Guillaumin, and Van~Gool]{food101}
Lukas Bossard, Matthieu Guillaumin, and Luc Van~Gool.
\newblock Food-101--mining discriminative components with random forests.
\newblock In \emph{Computer vision--ECCV 2014: 13th European conference, zurich, Switzerland, September 6-12, 2014, proceedings, part VI 13}, pages 446--461. Springer, 2014.

\bibitem[Cai and Vasconcelos(2018)]{cascade}
Zhaowei Cai and Nuno Vasconcelos.
\newblock Cascade r-cnn: Delving into high quality object detection.
\newblock In \emph{Proceedings of the IEEE conference on computer vision and pattern recognition}, pages 6154--6162, 2018.

\bibitem[Carion et~al.(2020)Carion, Massa, Synnaeve, Usunier, Kirillov, and Zagoruyko]{detr}
Nicolas Carion, Francisco Massa, Gabriel Synnaeve, Nicolas Usunier, Alexander Kirillov, and Sergey Zagoruyko.
\newblock End-to-end object detection with transformers.
\newblock In \emph{European conference on computer vision}, pages 213--229. Springer, 2020.

\bibitem[Caron et~al.(2021)Caron, Touvron, Misra, J{\'e}gou, Mairal, Bojanowski, and Joulin]{emerging}
Mathilde Caron, Hugo Touvron, Ishan Misra, Herv{\'e} J{\'e}gou, Julien Mairal, Piotr Bojanowski, and Armand Joulin.
\newblock Emerging properties in self-supervised vision transformers.
\newblock In \emph{Proceedings of the IEEE/CVF international conference on computer vision}, pages 9650--9660, 2021.

\bibitem[Chen et~al.(2019)Chen, Wang, Pang, Cao, Xiong, Li, Sun, Feng, Liu, Xu, Zhang, Cheng, Zhu, Cheng, Zhao, Li, Lu, Zhu, Wu, Dai, Wang, Shi, Ouyang, Loy, and Lin]{openmmlab}
Kai Chen, Jiaqi Wang, Jiangmiao Pang, Yuhang Cao, Yu Xiong, Xiaoxiao Li, Shuyang Sun, Wansen Feng, Ziwei Liu, Jiarui Xu, Zheng Zhang, Dazhi Cheng, Chenchen Zhu, Tianheng Cheng, Qijie Zhao, Buyu Li, Xin Lu, Rui Zhu, Yue Wu, Jifeng Dai, Jingdong Wang, Jianping Shi, Wanli Ouyang, Chen~Change Loy, and Dahua Lin.
\newblock Mmdetection: Open mmlab detection toolbox and benchmark, 2019.

\bibitem[Chen et~al.(2017)Chen, Papandreou, Kokkinos, Murphy, and Yuille]{chen2017deeplab}
Liang-Chieh Chen, George Papandreou, Iasonas Kokkinos, Kevin Murphy, and Alan~L Yuille.
\newblock Deeplab: Semantic image segmentation with deep convolutional nets, atrous convolution, and fully connected crfs.
\newblock \emph{IEEE transactions on pattern analysis and machine intelligence}, 40\penalty0 (4):\penalty0 834--848, 2017.

\bibitem[Chen et~al.(2022)Chen, Ge, Tong, Wang, Song, Wang, and Luo]{adaptformer}
Shoufa Chen, Chongjian Ge, Zhan Tong, Jiangliu Wang, Yibing Song, Jue Wang, and Ping Luo.
\newblock Adaptformer: Adapting vision transformers for scalable visual recognition.
\newblock \emph{Advances in Neural Information Processing Systems}, 35:\penalty0 16664--16678, 2022.

\bibitem[Chen and He(2021)]{chen2021exploring}
Xinlei Chen and Kaiming He.
\newblock Exploring simple siamese representation learning.
\newblock In \emph{Proceedings of the IEEE/CVF conference on computer vision and pattern recognition}, pages 15750--15758, 2021.

\bibitem[Chen et~al.(2021)Chen, Xie, and He]{empirical}
Xinlei Chen, Saining Xie, and Kaiming He.
\newblock An empirical study of training self-supervised vision transformers.
\newblock In \emph{Proceedings of the IEEE/CVF international conference on computer vision}, pages 9640--9649, 2021.

\bibitem[Cho et~al.(2024)Cho, Shin, Hong, Arnab, Seo, and Kim]{cho2024cat}
Seokju Cho, Heeseong Shin, Sunghwan Hong, Anurag Arnab, Paul~Hongsuck Seo, and Seungryong Kim.
\newblock Cat-seg: Cost aggregation for open-vocabulary semantic segmentation.
\newblock In \emph{Proceedings of the IEEE/CVF Conference on Computer Vision and Pattern Recognition}, pages 4113--4123, 2024.

\bibitem[Choi et~al.(2021)Choi, Lee, Kim, Kim, Kim, Sohn, and Min]{choi2021adaptive}
Hyesong Choi, Hunsang Lee, Sunkyung Kim, Sunok Kim, Seungryong Kim, Kwanghoon Sohn, and Dongbo Min.
\newblock Adaptive confidence thresholding for monocular depth estimation.
\newblock In \emph{Proceedings of the IEEE/CVF International Conference on Computer Vision}, pages 12808--12818, 2021.

\bibitem[Choi et~al.(2023{\natexlab{a}})Choi, Lee, Jeong, and Min]{ema2}
Hyesong Choi, Hunsang Lee, Seongwon Jeong, and Dongbo Min.
\newblock Environment agnostic representation for visual reinforcement learning.
\newblock In \emph{Proceedings of the IEEE/CVF International Conference on Computer Vision}, pages 263--273, 2023{\natexlab{a}}.

\bibitem[Choi et~al.(2023{\natexlab{b}})Choi, Lee, Song, Jeon, Sohn, and Min]{ema1}
Hyesong Choi, Hunsang Lee, Wonil Song, Sangryul Jeon, Kwanghoon Sohn, and Dongbo Min.
\newblock Local-guided global: Paired similarity representation for visual reinforcement learning.
\newblock In \emph{Proceedings of the IEEE/CVF Conference on Computer Vision and Pattern Recognition}, pages 15072--15082, 2023{\natexlab{b}}.

\bibitem[Choi et~al.(2024{\natexlab{a}})Choi, Lee, Joung, Park, Kim, and Min]{choi2024emerging}
Hyesong Choi, Hunsang Lee, Seyoung Joung, Hyejin Park, Jiyeong Kim, and Dongbo Min.
\newblock Emerging property of masked token for effective pre-training.
\newblock \emph{arXiv preprint arXiv:2404.08330}, 2024{\natexlab{a}}.

\bibitem[Choi et~al.(2024{\natexlab{b}})Choi, Park, Yi, Cha, and Min]{choi2024salience}
Hyesong Choi, Hyejin Park, Kwang~Moo Yi, Sungmin Cha, and Dongbo Min.
\newblock Salience-based adaptive masking: Revisiting token dynamics for enhanced pre-training.
\newblock \emph{arXiv preprint arXiv:2404.08327}, 2024{\natexlab{b}}.

\bibitem[Deng et~al.(2009)Deng, Dong, Socher, Li, Li, and Fei-Fei]{imagenet}
Jia Deng, Wei Dong, Richard Socher, Li-Jia Li, Kai Li, and Li Fei-Fei.
\newblock Imagenet: A large-scale hierarchical image database.
\newblock In \emph{2009 IEEE conference on computer vision and pattern recognition}, pages 248--255. Ieee, 2009.

\bibitem[Devlin(2018)]{bert}
Jacob Devlin.
\newblock Bert: Pre-training of deep bidirectional transformers for language understanding.
\newblock \emph{arXiv preprint arXiv:1810.04805}, 2018.

\bibitem[Dong et~al.(2015)Dong, Loy, He, and Tang]{dong2015image}
Chao Dong, Chen~Change Loy, Kaiming He, and Xiaoou Tang.
\newblock Image super-resolution using deep convolutional networks.
\newblock \emph{IEEE transactions on pattern analysis and machine intelligence}, 38\penalty0 (2):\penalty0 295--307, 2015.

\bibitem[Dosovitskiy et~al.(2021)Dosovitskiy, Beyer, Kolesnikov, Weissenborn, Zhai, Unterthiner, Dehghani, Minderer, Heigold, Gelly, Uszkoreit, and Houlsby]{vit}
Alexey Dosovitskiy, Lucas Beyer, Alexander Kolesnikov, Dirk Weissenborn, Xiaohua Zhai, Thomas Unterthiner, Mostafa Dehghani, Matthias Minderer, Georg Heigold, Sylvain Gelly, Jakob Uszkoreit, and Neil Houlsby.
\newblock An image is worth 16x16 words: Transformers for image recognition at scale, 2021.

\bibitem[Geiger et~al.(2012)Geiger, Lenz, and Urtasun]{kitti}
Andreas Geiger, Philip Lenz, and Raquel Urtasun.
\newblock Are we ready for autonomous driving? the kitti vision benchmark suite.
\newblock In \emph{2012 IEEE conference on computer vision and pattern recognition}, pages 3354--3361. IEEE, 2012.

\bibitem[Girshick et~al.(2014)Girshick, Donahue, Darrell, and Malik]{girshick2014rich}
Ross Girshick, Jeff Donahue, Trevor Darrell, and Jitendra Malik.
\newblock Rich feature hierarchies for accurate object detection and semantic segmentation.
\newblock In \emph{Proceedings of the IEEE conference on computer vision and pattern recognition}, pages 580--587, 2014.

\bibitem[Godard et~al.(2019)Godard, Mac~Aodha, Firman, and Brostow]{godard2019digging}
Cl{\'e}ment Godard, Oisin Mac~Aodha, Michael Firman, and Gabriel~J Brostow.
\newblock Digging into self-supervised monocular depth estimation.
\newblock In \emph{Proceedings of the IEEE/CVF international conference on computer vision}, pages 3828--3838, 2019.

\bibitem[Grill et~al.(2020)Grill, Strub, Altch{\'e}, Tallec, Richemond, Buchatskaya, Doersch, Avila~Pires, Guo, Gheshlaghi~Azar, et~al.]{grill2020bootstrap}
Jean-Bastien Grill, Florian Strub, Florent Altch{\'e}, Corentin Tallec, Pierre Richemond, Elena Buchatskaya, Carl Doersch, Bernardo Avila~Pires, Zhaohan Guo, Mohammad Gheshlaghi~Azar, et~al.
\newblock Bootstrap your own latent-a new approach to self-supervised learning.
\newblock \emph{Advances in neural information processing systems}, 33:\penalty0 21271--21284, 2020.

\bibitem[He et~al.(2023{\natexlab{a}})He, Cai, Zhang, Tao, and Zhuang]{spt}
Haoyu He, Jianfei Cai, Jing Zhang, Dacheng Tao, and Bohan Zhuang.
\newblock Sensitivity-aware visual parameter-efficient fine-tuning.
\newblock In \emph{Proceedings of the IEEE/CVF International Conference on Computer Vision}, pages 11825--11835, 2023{\natexlab{a}}.

\bibitem[He et~al.(2017)He, Gkioxari, Doll{\'a}r, and Girshick]{maskrcnn}
Kaiming He, Georgia Gkioxari, Piotr Doll{\'a}r, and Ross Girshick.
\newblock Mask r-cnn.
\newblock In \emph{Proceedings of the IEEE international conference on computer vision}, pages 2961--2969, 2017.

\bibitem[He et~al.(2022{\natexlab{a}})He, Chen, Xie, Li, Doll{\'a}r, and Girshick]{he2022masked}
Kaiming He, Xinlei Chen, Saining Xie, Yanghao Li, Piotr Doll{\'a}r, and Ross Girshick.
\newblock Masked autoencoders are scalable vision learners.
\newblock In \emph{Proceedings of the IEEE/CVF conference on computer vision and pattern recognition}, pages 16000--16009, 2022{\natexlab{a}}.

\bibitem[He et~al.(2022{\natexlab{b}})He, Chen, Xie, Li, Doll{\'a}r, and Girshick]{mae}
Kaiming He, Xinlei Chen, Saining Xie, Yanghao Li, Piotr Doll{\'a}r, and Ross Girshick.
\newblock Masked autoencoders are scalable vision learners.
\newblock In \emph{Proceedings of the IEEE/CVF conference on computer vision and pattern recognition}, pages 16000--16009, 2022{\natexlab{b}}.

\bibitem[He et~al.(2023{\natexlab{b}})He, Li, Zhang, Yang, and Wang]{kadaptation}
Xuehai He, Chunyuan Li, Pengchuan Zhang, Jianwei Yang, and Xin~Eric Wang.
\newblock Parameter-efficient model adaptation for vision transformers.
\newblock In \emph{Proceedings of the AAAI Conference on Artificial Intelligence}, pages 817--825, 2023{\natexlab{b}}.

\bibitem[Hong et~al.(2022)Hong, Cho, Nam, Lin, and Kim]{hong2022cost}
Sunghwan Hong, Seokju Cho, Jisu Nam, Stephen Lin, and Seungryong Kim.
\newblock Cost aggregation with 4d convolutional swin transformer for few-shot segmentation.
\newblock In \emph{European Conference on Computer Vision}, pages 108--126. Springer, 2022.

\bibitem[Houlsby et~al.(2019)Houlsby, Giurgiu, Jastrzebski, Morrone, De~Laroussilhe, Gesmundo, Attariyan, and Gelly]{adapter_nlp}
Neil Houlsby, Andrei Giurgiu, Stanislaw Jastrzebski, Bruna Morrone, Quentin De~Laroussilhe, Andrea Gesmundo, Mona Attariyan, and Sylvain Gelly.
\newblock Parameter-efficient transfer learning for nlp.
\newblock In \emph{International conference on machine learning}, pages 2790--2799. PMLR, 2019.

\bibitem[Hu et~al.(2021)Hu, Shen, Wallis, Allen-Zhu, Li, Wang, Wang, and Chen]{lora}
Edward~J Hu, Yelong Shen, Phillip Wallis, Zeyuan Allen-Zhu, Yuanzhi Li, Shean Wang, Lu Wang, and Weizhu Chen.
\newblock Lora: Low-rank adaptation of large language models.
\newblock \emph{arXiv preprint arXiv:2106.09685}, 2021.

\bibitem[Jia et~al.(2022)Jia, Tang, Chen, Cardie, Belongie, Hariharan, and Lim]{vpt}
Menglin Jia, Luming Tang, Bor-Chun Chen, Claire Cardie, Serge Belongie, Bharath Hariharan, and Ser-Nam Lim.
\newblock Visual prompt tuning.
\newblock In \emph{European Conference on Computer Vision}, pages 709--727. Springer, 2022.

\bibitem[Jia et~al.(2016)Jia, De~Brabandere, Tuytelaars, and Gool]{dfn}
Xu Jia, Bert De~Brabandere, Tinne Tuytelaars, and Luc~V Gool.
\newblock Dynamic filter networks.
\newblock \emph{Advances in neural information processing systems}, 29, 2016.

\bibitem[Kim et~al.(2022)Kim, Choi, and Min]{kim2022sequential}
Sunkyung Kim, Hyesong Choi, and Dongbo Min.
\newblock Sequential cross attention based multi-task learning.
\newblock In \emph{2022 IEEE International Conference on Image Processing (ICIP)}, pages 2311--2315. IEEE, 2022.

\bibitem[Krizhevsky et~al.(2009)Krizhevsky, Hinton, et~al.]{cifar100}
Alex Krizhevsky, Geoffrey Hinton, et~al.
\newblock Learning multiple layers of features from tiny images.
\newblock 2009.

\bibitem[Krizhevsky et~al.(2012)Krizhevsky, Sutskever, and Hinton]{krizhevsky2012imagenet}
Alex Krizhevsky, Ilya Sutskever, and Geoffrey~E Hinton.
\newblock Imagenet classification with deep convolutional neural networks.
\newblock \emph{Advances in neural information processing systems}, 25, 2012.

\bibitem[Lee et~al.(2022)Lee, Choi, Sohn, and Min]{lee2022knn}
Hunsang Lee, Hyesong Choi, Kwanghoon Sohn, and Dongbo Min.
\newblock Knn local attention for image restoration.
\newblock In \emph{Proceedings of the IEEE/CVF conference on computer vision and pattern recognition}, pages 2139--2149, 2022.

\bibitem[Lee et~al.(2023)Lee, Choi, Sohn, and Min]{lee2023cross}
Hunsang Lee, Hyesong Choi, Kwanghoon Sohn, and Dongbo Min.
\newblock Cross-scale knn image transformer for image restoration.
\newblock \emph{IEEE Access}, 11:\penalty0 13013--13027, 2023.

\bibitem[Lei et~al.(2023)Lei, Bai, Brahma, Ainslie, Lee, Zhou, Du, Zhao, Wu, Li, et~al.]{coda}
Tao Lei, Junwen Bai, Siddhartha Brahma, Joshua Ainslie, Kenton Lee, Yanqi Zhou, Nan Du, Vincent Zhao, Yuexin Wu, Bo Li, et~al.
\newblock Conditional adapters: Parameter-efficient transfer learning with fast inference.
\newblock \emph{Advances in Neural Information Processing Systems}, 36:\penalty0 8152--8172, 2023.

\bibitem[Lester et~al.(2021)Lester, Al-Rfou, and Constant]{prompt_nlp}
Brian Lester, Rami Al-Rfou, and Noah Constant.
\newblock The power of scale for parameter-efficient prompt tuning.
\newblock \emph{arXiv preprint arXiv:2104.08691}, 2021.

\bibitem[Li et~al.(2023)Li, Zhang, Xu, Liu, Zhang, Ni, and Shum]{mask_dino}
Feng Li, Hao Zhang, Huaizhe Xu, Shilong Liu, Lei Zhang, Lionel~M Ni, and Heung-Yeung Shum.
\newblock Mask dino: Towards a unified transformer-based framework for object detection and segmentation.
\newblock In \emph{Proceedings of the IEEE/CVF Conference on Computer Vision and Pattern Recognition}, pages 3041--3050, 2023.

\bibitem[Lin et~al.(2014)Lin, Maire, Belongie, Hays, Perona, Ramanan, Doll{\'a}r, and Zitnick]{coco}
Tsung-Yi Lin, Michael Maire, Serge Belongie, James Hays, Pietro Perona, Deva Ramanan, Piotr Doll{\'a}r, and C~Lawrence Zitnick.
\newblock Microsoft coco: Common objects in context.
\newblock In \emph{Computer Vision--ECCV 2014: 13th European Conference, Zurich, Switzerland, September 6-12, 2014, Proceedings, Part V 13}, pages 740--755. Springer, 2014.

\bibitem[Liu et~al.(2017)Liu, Breuel, and Kautz]{liu2017unsupervised}
Ming-Yu Liu, Thomas Breuel, and Jan Kautz.
\newblock Unsupervised image-to-image translation networks.
\newblock \emph{Advances in neural information processing systems}, 30, 2017.

\bibitem[Liu et~al.(2021)Liu, Lin, Cao, Hu, Wei, Zhang, Lin, and Guo]{swin}
Ze Liu, Yutong Lin, Yue Cao, Han Hu, Yixuan Wei, Zheng Zhang, Stephen Lin, and Baining Guo.
\newblock Swin transformer: Hierarchical vision transformer using shifted windows, 2021.

\bibitem[Liu et~al.(2022)Liu, Hu, Lin, Yao, Xie, Wei, Ning, Cao, Zhang, Dong, et~al.]{swinv2}
Ze Liu, Han Hu, Yutong Lin, Zhuliang Yao, Zhenda Xie, Yixuan Wei, Jia Ning, Yue Cao, Zheng Zhang, Li Dong, et~al.
\newblock Swin transformer v2: Scaling up capacity and resolution.
\newblock In \emph{Proceedings of the IEEE/CVF conference on computer vision and pattern recognition}, pages 12009--12019, 2022.

\bibitem[Long et~al.(2015)Long, Shelhamer, and Darrell]{long2015fully}
Jonathan Long, Evan Shelhamer, and Trevor Darrell.
\newblock Fully convolutional networks for semantic segmentation.
\newblock In \emph{Proceedings of the IEEE conference on computer vision and pattern recognition}, pages 3431--3440, 2015.

\bibitem[Netzer et~al.(2011)Netzer, Wang, Coates, Bissacco, Wu, Ng, et~al.]{svhn}
Yuval Netzer, Tao Wang, Adam Coates, Alessandro Bissacco, Baolin Wu, Andrew~Y Ng, et~al.
\newblock Reading digits in natural images with unsupervised feature learning.
\newblock In \emph{NIPS workshop on deep learning and unsupervised feature learning}, page~4. Granada, 2011.

\bibitem[Qi et~al.(2022)Qi, Ruan, Zuo, and Li]{LN-tuning}
Wang Qi, Yu-Ping Ruan, Yuan Zuo, and Taihao Li.
\newblock Parameter-efficient tuning on layer normalization for pre-trained language models, 2022.

\bibitem[Ranftl et~al.(2021)Ranftl, Bochkovskiy, and Koltun]{dpt}
Ren{\'e} Ranftl, Alexey Bochkovskiy, and Vladlen Koltun.
\newblock Vision transformers for dense prediction.
\newblock In \emph{Proceedings of the IEEE/CVF international conference on computer vision}, pages 12179--12188, 2021.

\bibitem[Redmon et~al.(2016)Redmon, Divvala, Girshick, and Farhadi]{redmon2016you}
Joseph Redmon, Santosh Divvala, Ross Girshick, and Ali Farhadi.
\newblock You only look once: Unified, real-time object detection.
\newblock In \emph{Proceedings of the IEEE conference on computer vision and pattern recognition}, pages 779--788, 2016.

\bibitem[Ren et~al.(2016)Ren, He, Girshick, and Sun]{ren2016faster}
Shaoqing Ren, Kaiming He, Ross Girshick, and Jian Sun.
\newblock Faster r-cnn: Towards real-time object detection with region proposal networks.
\newblock \emph{IEEE transactions on pattern analysis and machine intelligence}, 39\penalty0 (6):\penalty0 1137--1149, 2016.

\bibitem[Silberman et~al.(2012)Silberman, Hoiem, Kohli, and Fergus]{nyuv2}
Nathan Silberman, Derek Hoiem, Pushmeet Kohli, and Rob Fergus.
\newblock Indoor segmentation and support inference from rgbd images.
\newblock In \emph{Computer Vision--ECCV 2012: 12th European Conference on Computer Vision, Florence, Italy, October 7-13, 2012, Proceedings, Part V 12}, pages 746--760. Springer, 2012.

\bibitem[Vaswani(2017)]{transformer}
Ashish Vaswani.
\newblock Attention is all you need.
\newblock \emph{arXiv preprint arXiv:1706.03762}, 2017.

\bibitem[Wang et~al.(2021)Wang, Xie, Li, Fan, Song, Liang, Lu, Luo, and Shao]{pvt}
Wenhai Wang, Enze Xie, Xiang Li, Deng-Ping Fan, Kaitao Song, Ding Liang, Tong Lu, Ping Luo, and Ling Shao.
\newblock Pyramid vision transformer: A versatile backbone for dense prediction without convolutions.
\newblock In \emph{Proceedings of the IEEE/CVF international conference on computer vision}, pages 568--578, 2021.

\bibitem[Xia et~al.(2024)Xia, Wang, Lv, Hao, and Shi]{vit-comer}
Chunlong Xia, Xinliang Wang, Feng Lv, Xin Hao, and Yifeng Shi.
\newblock Vit-comer: Vision transformer with convolutional multi-scale feature interaction for dense predictions.
\newblock In \emph{Proceedings of the IEEE/CVF Conference on Computer Vision and Pattern Recognition}, pages 5493--5502, 2024.

\bibitem[Xiao et~al.(2018)Xiao, Liu, Zhou, Jiang, and Sun]{upernet}
Tete Xiao, Yingcheng Liu, Bolei Zhou, Yuning Jiang, and Jian Sun.
\newblock Unified perceptual parsing for scene understanding.
\newblock In \emph{Proceedings of the European conference on computer vision (ECCV)}, pages 418--434, 2018.

\bibitem[Xie et~al.(2021)Xie, Wang, Yu, Anandkumar, Alvarez, and Luo]{segformer}
Enze Xie, Wenhai Wang, Zhiding Yu, Anima Anandkumar, Jose~M Alvarez, and Ping Luo.
\newblock Segformer: Simple and efficient design for semantic segmentation with transformers.
\newblock \emph{Advances in neural information processing systems}, 34:\penalty0 12077--12090, 2021.

\bibitem[Xie et~al.(2022{\natexlab{a}})Xie, Geng, Hu, Zhang, Hu, and Cao]{mim}
Zhenda Xie, Zigang Geng, Jingcheng Hu, Zheng Zhang, Han Hu, and Yue Cao.
\newblock Revealing the dark secrets of masked image modeling, 2022{\natexlab{a}}.

\bibitem[Xie et~al.(2022{\natexlab{b}})Xie, Zhang, Cao, Lin, Bao, Yao, Dai, and Hu]{simmim}
Zhenda Xie, Zheng Zhang, Yue Cao, Yutong Lin, Jianmin Bao, Zhuliang Yao, Qi Dai, and Han Hu.
\newblock Simmim: A simple framework for masked image modeling.
\newblock In \emph{Proceedings of the IEEE/CVF conference on computer vision and pattern recognition}, pages 9653--9663, 2022{\natexlab{b}}.

\bibitem[Xie et~al.(2022{\natexlab{c}})Xie, Zhang, Cao, Lin, Bao, Yao, Dai, and Hu]{xie2022simmim}
Zhenda Xie, Zheng Zhang, Yue Cao, Yutong Lin, Jianmin Bao, Zhuliang Yao, Qi Dai, and Han Hu.
\newblock Simmim: A simple framework for masked image modeling.
\newblock In \emph{Proceedings of the IEEE/CVF conference on computer vision and pattern recognition}, pages 9653--9663, 2022{\natexlab{c}}.

\bibitem[Yi et~al.(2022)Yi, Ge, Li, Yang, Li, Wu, Shan, and Qie]{yi2022masked}
Kun Yi, Yixiao Ge, Xiaotong Li, Shusheng Yang, Dian Li, Jianping Wu, Ying Shan, and Xiaohu Qie.
\newblock Masked image modeling with denoising contrast.
\newblock \emph{arXiv preprint arXiv:2205.09616}, 2022.

\bibitem[Yin et~al.(2023)Yin, Yang, Wang, Yu, Wei, and Sun]{lorand}
Dongshuo Yin, Yiran Yang, Zhechao Wang, Hongfeng Yu, Kaiwen Wei, and Xian Sun.
\newblock 1\% vs 100\%: Parameter-efficient low rank adapter for dense predictions.
\newblock In \emph{Proceedings of the IEEE/CVF Conference on Computer Vision and Pattern Recognition}, pages 20116--20126, 2023.

\bibitem[Yosinski et~al.(2014)Yosinski, Clune, Bengio, and Lipson]{partial}
Jason Yosinski, Jeff Clune, Yoshua Bengio, and Hod Lipson.
\newblock How transferable are features in deep neural networks?, 2014.

\bibitem[Zaken et~al.(2021)Zaken, Ravfogel, and Goldberg]{bitfit}
Elad~Ben Zaken, Shauli Ravfogel, and Yoav Goldberg.
\newblock Bitfit: Simple parameter-efficient fine-tuning for transformer-based masked language-models.
\newblock \emph{arXiv preprint arXiv:2106.10199}, 2021.

\bibitem[Zhou et~al.(2017)Zhou, Zhao, Puig, Fidler, Barriuso, and Torralba]{ade20k}
Bolei Zhou, Hang Zhao, Xavier Puig, Sanja Fidler, Adela Barriuso, and Antonio Torralba.
\newblock Scene parsing through ade20k dataset.
\newblock In \emph{Proceedings of the IEEE conference on computer vision and pattern recognition}, pages 633--641, 2017.

\end{thebibliography}
}

\end{document}